\documentclass[letterpaper]{article} 
\usepackage[preprint]{aaai2027}  
\usepackage[hyphens]{url}  
\usepackage{graphicx} 
\usepackage{natbib}  
\usepackage{caption} 
\usepackage{algorithm}
\usepackage{algorithmic}
\usepackage{xspace}
\usepackage[table]{xcolor}
\usepackage{newfloat}
\usepackage{listings}
\DeclareCaptionStyle{ruled}{labelfont=normalfont,labelsep=colon,strut=off} 
\floatstyle{ruled}
\newfloat{listing}{tb}{lst}{}
\floatname{listing}{Listing}

\usepackage{booktabs}
\usepackage{amsfonts}
\usepackage{amsmath}
\usepackage{soul}
\title{Video Compression with Graph-inspired Neural Representation}

\author{
    Changqi Wang\textsuperscript{\rm 1}, Ge Gao\textsuperscript{\rm 1}, Fan Zhang\textsuperscript{\rm 1}, Yue Li\textsuperscript{\rm 2}, Kai Zhang\textsuperscript{\rm 2}, Li Zhang\textsuperscript{\rm 2}, David Bull\textsuperscript{\rm 1} 
}
\affiliations{
    \textsuperscript{\rm 1}University of Bristol\\
    \textsuperscript{\rm 2}Bytedance Inc., San Diego, USA\\

    \{changqi.wang, ge1.gao, fan.zhang, dave.bull\}@bristol.ac.uk\\
    \{yue.li, zhangkai.video, lizhang.idm\}@bytedance.com
}

\newcommand{\green}{\textcolor{green!40!black!50}}
\newcommand{\ie}{\textit{i.e.,}}

\newcommand{\name}{G-NeRV\xspace}

\begin{document}

\maketitle

\begin{abstract}
Implicit Neural Representations (INR) provide a compact and content-adaptive paradigm for video compression, typically representing a video through shared network parameters and frame-indexed embeddings. Compared to conventional or autoencoder-based codecs, these approaches exploit temporal redundancy within videos in an implicit manner, which potentially results in sub-optimal compression performance. In this paper, we propose \name, a graph-inspired INR that explicitly improves temporal redundancy exploitation in the implicit latent space. Motivated by the total correlation principles in information theory, we construct a temporal neighborhood over frame embeddings and perform message passing to aggregate reusable information from neighboring frames through an adaptive gate controlling the injection of neighboring information. Inspired by the reference frame buffer in conventional video coding, a memory bank mechanism has been further designed to enable efficient temporal-neighbor retrieval under random frame-index sampling in INR training. This new representation model has been integrated into an advanced representation compression framework and compared with existing conventional and neural video codecs. The results show that the \name codec outperforms the state-of-the-art INR-based codec, NVRC, and the latest standard video codec, VVC VTM, by 8.86\% and 14.68\% (in BD-rate), respectively, measured by PSNR on the UVG dataset. 
\end{abstract}


\section{Introduction}

Video compression is a key technique for transmitting and storing the ever-growing volume of video data, with the goal of compressing video sequences into compact bitstreams while maintaining reconstruction quality.
Conventional codecs such as HEVC HM~\cite{HEVC_overview} and VVC VTM~\cite{VVC_overview} adopt a hybrid coding pipeline comprising motion prediction, residual transform, quantization, and entropy coding, which has been predominant in practical applications. Recently, neural video codecs have been proposed~\cite{DVC,DCVC}, which typically employ end-to-end optimized autoencoder networks to replace hand-designed modules in conventional codecs to achieve efficient representation in the feature domain. Although these learning-based video codecs have achieved competitive compression performance, most of them use fixed, offline-optimized network parameters during inference, which often makes them sensitive to cross-domain distribution shifts~\cite{adaptive1,adaptive2} and results in relatively high decoding complexity due to the large network capacity required for model generalization~\cite{Review_Gao}.

To address these issues, implicit neural representations (INRs)~\cite{INR_NeRF} have been applied in compact video representation and compression~\cite{NeRV,HNeRV}, following their success in 3D reconstruction. These INR-based video codecs typically overfit a lightweight network for a specific video instance, which is naturally content-adaptive and has low decoding complexity. The latest works in this area~\cite{NVRC} have reported promising results when compared to standard and autoencoder-based video codecs, and there are further contributions focusing on their practical deployment in terms of low latency \cite{PNVC,GIViC} and high scalability~\cite{NVRC++}.

\begin{figure}[t]
    \centering
    \includegraphics[width=1.0\linewidth]{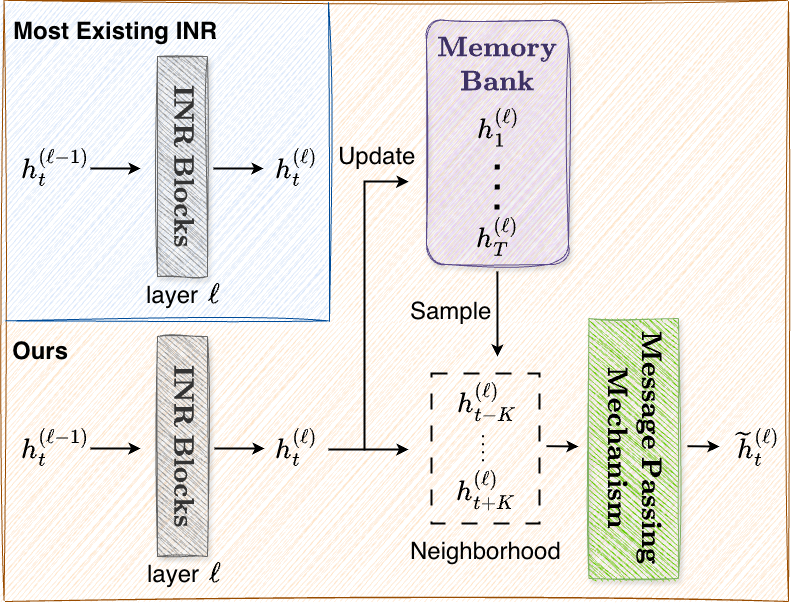}
    \caption{Blue background: most existing INR frameworks; Orange background: message passing mechanism and memory bank in \name. $h$ denotes the latent embedding.}
    \label{fig:cover}
\end{figure}

It is noted that, due to the unique representation process within INR-based video codecs, the effective exploitation of temporal redundancy, which is key for video coding, has not been fully investigated in the literature. Different from conventional and autoencoder-based codecs, where temporal redundancy is exploited explicitly through motion prediction and conditional learning, INR-based codecs, in particular those based on different neural representations for videos (NeRVs)~\cite{NeRV,HiNeRV}, learn the temporal correlation between frames in an implicit way through network overfitting. It is not clear whether this implicit exploitation is sufficient compared to explicit methods.

To address this research gap, this paper first studies the exploitation of temporal redundancy in NeRV-based codecs from the perspective of information theory. Based on this study and inspired by the advances in graph neural networks \cite{GIN,GCN}, we propose, \textbf{\name}, a graph-inspired neural representation framework for video compression. Specifically, we define a \emph{temporal neighborhood} for frame embeddings with INRs and design a \emph{message passing mechanism} to aggregate reusable information from neighboring embeddings. To prevent aggregating frame-specific variations during indiscriminative aggregation, we further introduce a learnable \emph{adaptive gate} in this mechanism to control the sharing of neighboring information. Moreover, to address the availability issue of temporal neighbors due to the random frame-index sampling in NeRVs, we also adopt a \emph{memory bank}, motivated by the reference buffer in conventional codecs, to enable efficient temporal neighbor retrieval during training. The primary contributions of this work are summarized below.

\begin{itemize}
    \item \name is the \textbf{first graph-inspired INR}, which enhances the exploitation of temporal redundancy in INR video codecs.
    \item \name employs a \textbf{temporal neighborhood} for embeddings in the latent space and improves temporal information sharing through \textbf{message passing mechanism}, rather than learning temporal correlation implicitly in many existing INR-based codecs (as shown in Figure \ref{fig:cover}).
    \item \name also integrates a \textbf{memory bank} to store neighboring embeddings during training, which \textbf{enables neighborhood construction} under random frame-index sampling in most INR-based codecs.
\end{itemize}

Our proposed \name has been integrated with an advanced neural video representation compression framework, NVRC \cite{NVRC}, to enable quantization, entropy coding, and end-to-end optimization. The resulting codec has been evaluated on three commonly used datasets and benchmarked against two conventional, two autoencoder-based, and three INR-based video codecs. The results demonstrate the superior performance of \name, with 14.68\%, 32.73\%, and 8.86\% coding gains compared to VTM 20.0 (RA), DCVC-RT, and NVRC, respectively, on the UVG database. It also outperforms the temporal-aware INR-based codec, DNeRV~\cite{DNeRV}, by 7.91\% in BD-rate.

\section{Related Work}

\subsection{Neural Video Compression}
Compared to conventional video codecs~\cite{HEVC_overview,VVC_overview} that are based on hand-crafted coding tools, neural video compression jointly optimizes learnable transforms, prediction, and entropy modeling networks in an end-to-end manner. This direction was pioneered by DVC~\cite{DVC} and further advanced by the DCVC family~\cite{DCVC,DCVC_DC,DCVC-UF} and other autoencoder-based codecs~\cite{ECVC,Hierarchical_autoencoder}. These neural codecs typically employ a network with a relatively large model capacity that is pre-trained for online inference. This potentially results in sub-optimal performance for unseen content and high decoding complexity. In contrast, INR-based codecs provide a sequence-adaptive alternative by representing videos through overfitted neural functions and associated latent parameters~\cite{UARNVC,NIRVANA,EntropyConstrained,NVRC-lite}. One notable family, including NeRV and its variants~\cite{HNeRV,SNeRV}, encodes a video sequence using frame/patch index conditioned neural representations with shared decoder parameters. Another group of methods, represented by COOL-CHIC-video~\cite{COOL-CHIC-video}, follows an overfitted coding paradigm with lightweight learnable latents and a compact synthesis network, and has been further improved by more efficient parameter allocation and coding tools~\cite{CNVC,Improved_COOL-CHIC}.

\subsection{Temporal Redundancy Exploitation}
In INR-based codecs, the exploitation of temporal redundancy varies across different INR paradigms. COOL-CHIC-based methods introduce explicit inter-coding mechanisms, as decoded reference-frame signals provide pixel-domain operands for motion-compensated prediction~\cite{COOL-CHIC-video,CNVC}.
In NeRV codecs, however, the coded information is primarily distributed across shared network parameters and frame-indexed embeddings, rather than organized as explicit per-frame prediction residuals. As a result, temporal prediction or pixel-domain warping cannot be directly applied in the latent domain. Some NeRV-based methods, therefore, exploit temporal redundancy indirectly, such as augmenting the model with motion or frame differences~\cite{FFNeRV,DNeRV}, adopting sequence-adaptive sampling strategies~\cite{BoostingNeRV,TreeNeRV,DSNeRV,TeNeRV}, or designing INR architectures with temporal modeling biases~\cite{TNeRV,CWRNN,HANeRV}. However, these methods have been reported to still be outperformed by the most advanced INR-based codecs \cite{NVRC} and conventional codecs~\cite{VVC_overview}.

\subsection{Graph Message Passing}
Graph Neural Networks (GNNs) operate on graph-structured data through message passing, where each node aggregates information from its neighbors according to the graph adjacency and updates its own embeddings. Various message-passing architectures~\cite{GAT,DeepGCNs,DGCNN} have been developed to capture structural dependencies, and graph-based or graph-inspired models have been applied to recent low-level vision tasks such as image denoising, resolution, and compression~\cite{Graph_denoising,Graph_resolution,Graph_compression} to aggregate useful information from spatial neighbors. However, graph-inspired message passing has rarely been explored for temporal redundancy exploitation in INR-based video compression at the embedding level.

\section{Methodology}

\subsection{Preliminaries}
\paragraph{Neural Video Representation.}
In an advanced neural video representation model~\cite{HiNeRV}, it represents a video sequence $\mathcal{X}=\{x_t\}_{t=1}^{T}$ with $T$ frames, where $x_t\in\mathbb{R}^{C\times H\times W}$. using a network $f_\theta$ with learnable parameters $\theta$. For generality, we absorb all learnable variables in $\theta$, including the index encoding module which encodes the temporal index $t$ to high-dimensional embedding. $f_\theta$ also contains a few INR blocks, each of which produces embeddings at different layers. 

\paragraph{Temporal Correlation of NeRV Embeddings.} In information theory, for a collection of random variables, total correlation
(TC)~\cite{TC} quantifies their collective statistical dependence and vanishes if and only if they are mutually independent.
In neural video representation, let $\mathcal{X}$ be a random video drawn from the video distribution, and $h_{t}^{(\ell)}$ denote the corresponding layer-$\ell$ embedding at time $t$. Without loss of generality, we drop the superscript $\ell$ below.
Here, we define the temporal neighborhood centered at $t$ as:
\begin{equation}
    \mathbf{H}_{t} =\left\{h_{t-K},\cdots,h_{t+K}\right\},
\end{equation}
in which $K$ is the radius of the neighborhood, and $2K+1$ is the size. Its total correlation is:
\begin{equation}
\begin{aligned}
    \mathrm{TC}(\mathbf{H}_{t})&=D_\mathrm{KL}\left(P_{\mathbf{H}_{t}}
        \|\prod_{j=-K}^{K}P_{{h}_{t+j}}\right) \\
    &=
    \sum_{j=-K}^{K}
        \mathcal{H}(h_{t+j})-\mathcal{H}({\mathbf{H}}_{t}),
\end{aligned}
\end{equation}
where $\mathcal{H}(\cdot)$ denotes Shannon entropy, and $\mathcal{H}({\mathbf{H}}_{t})$ stands for the joint entropy of the random variables in the neighborhood. $D_\mathrm{KL}$ represents the KL divergence, 

\paragraph{Motivation.}
We aim to exploit temporal redundancy by modeling the TC among frame-indexed embeddings in the INR latent space. For analysis, we conceptually decompose the intermediate embedding at layer $\ell$ through simple residual learning:
\begin{equation}
    h_t = s_t + r_t,
\end{equation}
in which $s_t$ denotes the temporally reusable component shared or predictable across neighboring frames, and $r_t$ stands for the frame-specific component required for motion, occlusion, and deformation. Given the center index $t$ and temporal radius $K$, we define the effective temporal TC as:
\begin{equation}
\begin{aligned}
    \mathrm{TC}_{\mathrm{eff}}
    &=\sum_{j=-K}^{K} \mathcal{H}\left(s_{t+j}\right)
    -\mathcal{H}\left(s_{t-K},\cdots,s_{t+K}\right) \\
    &\simeq L_{\mathrm{ind}} - L_{\mathrm{joint}}.
\end{aligned}
\end{equation}
Here $L_{\mathrm{ind}}$ and $L_{\mathrm{joint}}$ are the description lengths of independently and jointly representing the reusable embedding components \cite{DescriptionLength}, respectively.
Similarly, we define the ineffective temporal TC as:
\begin{equation}
    \mathrm{TC}_{\mathrm{ineff}}
    =\sum_{j=-K}^{K} \mathcal{H}\left(r_{t+j}\right)
    -\mathcal{H}\left(r_{t-K},\cdots,r_{t+K}\right)
\end{equation}

A larger $\mathrm{TC}_{\mathrm{eff}}$ therefore indicates a smaller description length for jointly representing the reusable embedding components, which means a greater opportunity to reuse shared temporal information, potentially reducing the required coding rate under the same reconstruction fidelity or improving reconstruction quality under a fixed rate budget. However, temporal dependence should not be maximized indiscriminately, as excessive coupling of the frame-specific component $r_t$ may suppress temporal differences and cause over-smoothed motion. 

Our design, therefore, \textbf{enhances effective temporal TC} through graph-inspired message passing over local temporal neighboring embeddings and \textbf{constrains ineffective temporal TC} with an adaptive gate that controls neighbor-information aggregation. Consequently, our framework serves as an embedding-level temporal redundancy exploitation mechanism for NeRV-based codecs, leveraging controlled temporal information sharing among frame-indexed embeddings.

\begin{figure*}[t]
    \centering
    \includegraphics[width=1\linewidth]{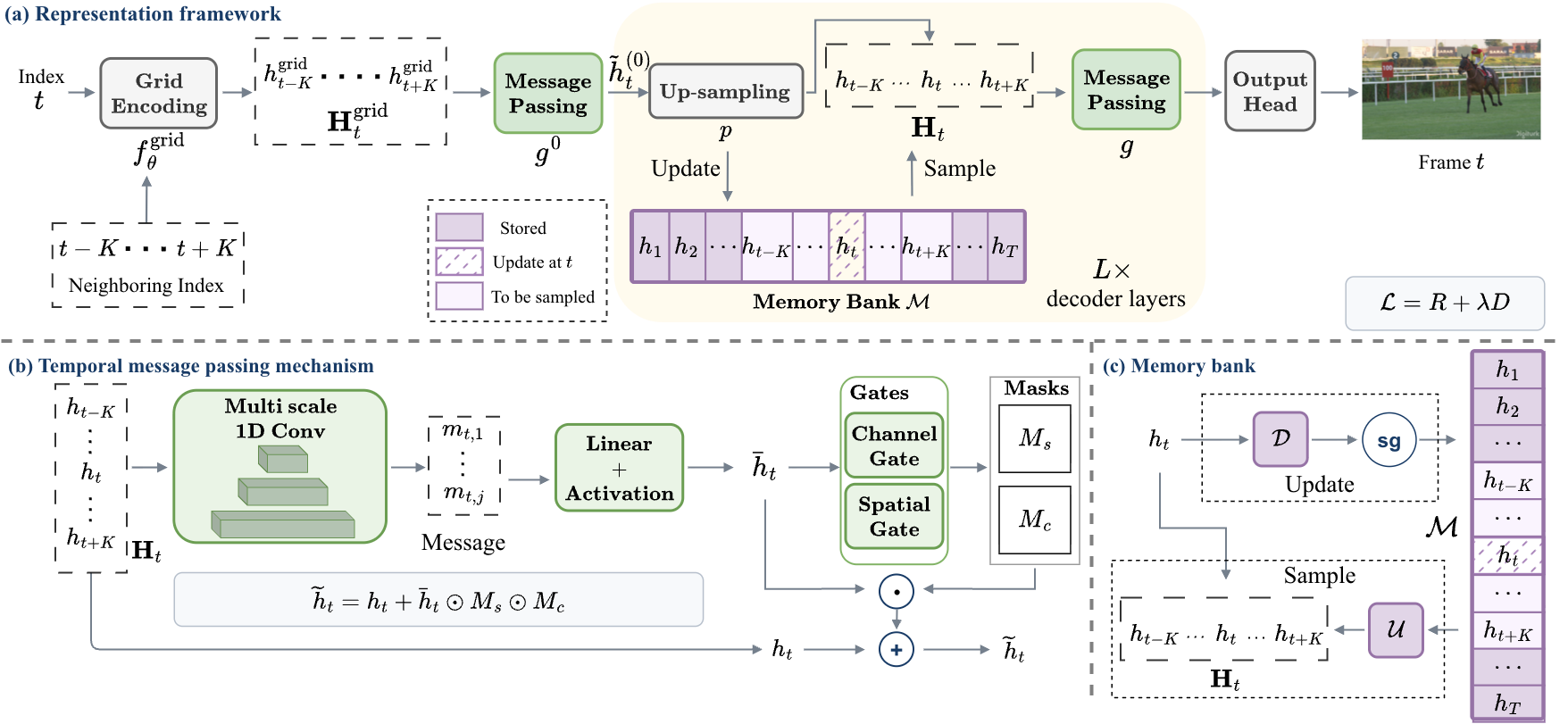}
    \caption {The illustration of the proposed \name framework}
    \label{fig:Framework}
\end{figure*}

\subsection{Overall Framework}

The proposed \name framework is illustrated in Figure ~\ref{fig:Framework}, which follows a typical representation pipeline described above. The primary innovation involves (i) a new graph-inspired message passing mechanism after grid encoding and each decoding block, and (ii) a memory bank that shares neighboring embeddings to enhance the message passing within each decoding block.

Specifically, the representation network $f_\theta$ in our framework is composed of grid-based encoding ($f_\theta^{\text{grid}}$) and a multi-layer decoder ($\{f^{(\ell)}_\theta\}_{\ell=1}^L$). The former maps a frame index $t$ to an initial grid embedding, \ie $h_t^{\text{grid}}=f_\theta^{\text{grid}}(t)$. To inject temporal information during grid-based encoding, we construct a local temporal neighborhood around $t$ by taking $\{t-K, \cdots, t+K\}$ as the input for $f^{\text{grid}}_\theta$:
\begin{equation}
    \mathbf{H}_t^{\text{grid}}=\left\{h_{t-K}^{\text{grid}},\cdots,h_{t+K}^{\text{grid}}\right\},
\end{equation}
where \textbf{out-of-range indices are clipped} to $[1,T]$.
This neighborhood is fed into the graph-inspired message passing module $g^{(0)}$ for grid encoding to obtain the temporally enhanced center embedding:
\begin{equation}
    \widetilde{h}_t^{(0)}=g^{(0)}(\mathbf{H}_t^{\text{grid}}).
\end{equation}

The embedding output $\widetilde{h}_t^{(0)}$ is then decoded via $L$ convolution-based decoder blocks $\{f^{(\ell)}_\theta\}_{\ell=1}^L$. Each decoder block contains an up-sampling block $p(\cdot)$ and a graph-inspired message passing module $g(\cdot)$ (with the same network architecture of $g^{(0)}$. In order to retrieve local temporal neighbors for message passing, we further construct a memory bank for neighbor embedding accessibility during training. The output of the final decoder block will then be passed through an output head to obtain the reconstructed frame $\hat{x}_t$.

\subsection{Temporal Message Passing Mechanism}\label{sec:message passing}

Inspired by advances in graph neural networks \cite{GraphReview}, The temporal message passing mechanism proposed in \name, considers each frame-level embedding as a node. Their temporal correlation is learned here to enhance the representation ability of the framework through this mechanism. As mentioned above, two primary issues are addressed here: (i) enhancing effective temporal TC and (ii) constraining ineffective TC among the embeddings in the neighborhood.

During grid-based encoding or at each decoding block, without loss of generality, the message passing module takes the temporal neighborhood $\mathbf{H}_{t}$, and employs $J$ one-dimensional (along the temporal axis) depth-wise convolutional blocks $u(\cdot)$ with different scales, \ie $\{u_j(\cdot)\}_{j=1}^{J}$. Here, various scales at each block correspond to different temporal receptive fields.
This design allows the module to capture both short-range frame-to-frame correlations and longer-range temporal dependencies in the neighborhood. Each convolutional block here processes the same embedding neighborhood input but only outputs a single frame embedding, \textit{the message} $m_{t,j}$, corresponding to frame $t$. A total of $J$ messages from these multi-scale convolutional layers are then concatenated and mixed by a channel mixer consisting of two linear layers $W_1$ and $W_2$, and a nonlinear activation $\sigma$ to obtain the aggregated temporal message $\bar{h}_{t}$:
\begin{equation}\label{eq:mp_details}
    \bar{h}_{t}=W_2\sigma\left(W_1\operatorname{Concat}\left(m_{t,1},\cdots,m_{t,J}\right)\right).
\end{equation}

To avoid indiscriminate temporal aggregation, we introduce an adaptive gate, which contains two lightweight convolution-based predictors, $\delta_s(\cdot)$ and $\delta_c(\cdot)$, with sigmoid functions used to estimate spatial- and channel-wise confidence masks:
\begin{equation}
    M_s=\delta_s(\bar{h}_t),\ \text{and} \ M_c=\delta_c(\bar{h}_t).
\end{equation}
Here, large gate values allow reusable temporal messages to be injected, while small gate values suppress unreliable neighbor information and preserve frame-specific details. These two confidence maps are re-sampled to the same resolution of $\bar{h}_t$ and are used to control the contribution of neighboring information. Finally, we use a residual connection to preserve the original frame-specific representation and alleviate the over-smoothing problem in GNNs~\cite{DeepGCNs}. The entire temporal message passing process is formulated as:
\begin{equation}\label{eq:mp_iter}
\begin{aligned}
    \widetilde{h}_t&=g(\mathbf{H}_t)
    =h_t+\bar{h}_{t}\odot M_s\odot M_c.
\end{aligned}
\end{equation}

\subsection{Memory Bank}

While integrating the temporal message passing mechanism into the representation framework can potentially improve the exploitation of temporal redundancy, constructing the temporal neighborhood at each decoding layer is extremely computationally expensive, particularly under random frame-index sampling \cite{NeRV}, where neighboring embeddings are not available when decoding the current frame. We therefore introduce a memory bank for each decoder block\footnote{We do not use a memory bank for the message passing module during grid encoding due to the minimal computational cost for obtaining initial grid embeddings there.} to store all the intermediate embeddings of the video sequence during training and retrieve temporal neighbors when constructing the neighborhood. Due to their large memory footprint, the neighborhood embeddings in the memory bank should not be transmitted within the bitstream; instead, they should be reconstructed at the decoder, inspired by the reference buffer used in modern conventional codecs~\cite{ReferenceBuffer}. Meanwhile, to further reduce GPU memory consumption, we only store low-resolution embeddings and perform learnable up-sampling to restore their full-resolution counterparts.

\paragraph{Sampling Strategy.}
To reduce memory consumption, we store down-scaled versions of embeddings in the memory bank rather than their full resolution counterparts. Specifically, each input embedding $h_t$ is first down-sampled by a $1\times1$ convolution block along the channel dimension and then spatially down-sampled via average pooling:
\begin{equation}
    h_t^{(\text{low})}=\mathcal{D}(h_t)=\operatorname{AvgPool}\left(\operatorname{Conv}(h_t)\right),
\end{equation}
where $h_t^{(\text{low})}$ is stored in the memory bank.
The neighbor embeddings in the temporal neighborhood $\mathbf{H}_t$ are constructed by sampling low-resolution embeddings from the memory bank and then restoring them to full resolution for the message passing process using a up-sampling module $\mathcal{U}$ which consists of nearest-neighbor up-sampler and a $1\times1$ convolution block.

\paragraph{Update Strategy.}
At the training stage, the goals of our design are efficient training and practical computation. For each selected decoder block, we first randomly initialize a memory bank $\mathcal{M}_t$ indexed by the current frame. With an embedding $h_t$ output by the up-sampling block $p(\cdot)$ in this decoder block, we update the corresponding memory entry by
\begin{equation}
    \mathcal{M}_t
    \leftarrow\operatorname{sg}\left(\mathcal{D}(h_t)\right),
\end{equation}
where $\operatorname{sg}(\cdot)$ denotes stop-gradient and $\mathcal{D}(\cdot)$ is the down-sampling operator.
During training, the memory bank is progressively updated and becomes increasingly informative.

In the decoding stage, the goals of our design are accurate reconstruction and fast decoding.
The former means that we need to keep all the same entries in the memory bank and modules in the representation framework, especially the down-sampling module $\mathcal{D}$ and up-sampling module $\mathcal{U}$ as those in the training stage.
The latter means that repeated computation should be avoided when conducting neighborhood.
Therefore, different from the sampling strategy used in the training process, decoding proceeds with index-order sweeping to fill the memory bank from the first entry to the last entry and constructs an auxiliary queue $\mathcal{Q}$ to avoid repeatedly computing embeddings. At the beginning, both the memory bank and the queue are set to be empty. To obtain $\widetilde{h}_t$, we first compute $h_t$ and the aggregated temporal message $\bar{h}_t$ using Equation~(\ref{eq:mp_details}), which requires the neighborhood $H_t$ and the calculation of $\{h_i\}_{i=t+1}^{t+K}$. We will then store the full-resolution embedding $\{h_i\}_{i=t+1}^{t+K}$ into the queue $\mathcal{Q}$ to avoid a second calculation when obtaining the $\{\widetilde{h}_i\}_{i=t+1}^{t+K}$, and store the low-resolution version of $\{h_i\}_{i=t+1}^{t+K}$ into the memory bank for a same neighborhood construction as that in the training.
The size of the queue is related to the size of the temporal neighborhood and is much smaller than the size of the memory bank - this keeps the computation practical.

\subsection{\name Video Codec}

To evaluate the effectiveness of the proposed \name representation framework, we integrate it into an advanced neural representation compression model, NVRC~\cite{NVRC}.
We implemented the same grid-based encoding $f^{\text{grid}}_\theta$, and up-sampling blocks $p(\cdot)$ as those in NVRC (also the same as in HiNeRV \cite{HiNeRV}).
The representation network $f_\theta$ together with the proposed temporal message passing blocks and memory bank, is compressed into the bitstream by learnable quantization and entropy models in NVRC. The entire codec is optimized in an end-to-end manner with a single rate-distortion objective, where the rate $R$ is measured as the sum of the quantized  and entropy coded learnable parameters in INR, quantization models, and entropy models. It is noted that the memory bank is dynamically generated through the encoding and decoding process, so only its associated trainable parameters (i.e., $\mathcal{D}$ and $\mathcal{U}$) are compressed alongside other INR parameters for transmission.
.

The distortion is measured between the reconstructed and original frames, \ie $D=\frac{1}{T}\sum_{t=1}^{T}d(x_t,\hat{x}_t)$, where $d(\cdot)$ is a combination of $\ell_2$ distance and MS-SSIM loss, following NVRC.
The total loss objective for end-to-end optimization is:
\begin{equation}
    \mathcal{L}=R+\lambda D,
\end{equation}
where $\lambda$ controls the rate-distortion trade-off.

\section{Experiments}

\subsection{Implementation Details}
Similar to other NeRV-like INR-based codecs, we configured \name with four different complexity scales for 4 different rates by using different channel dimensions in the decoder latent space and feature grids.
For settings related to the module designs in our framework, we set the radius of the neighborhood $K=3$.
The ratios for channel downscale and spatial downscale are $2$ and $4$, respectively, in the downsampling $\mathcal{D}$ in memory bank construction.
The general settings can be seen in the Appendix.

\begin{figure*}[t]
    \centering
    \includegraphics[width=1.0\linewidth]{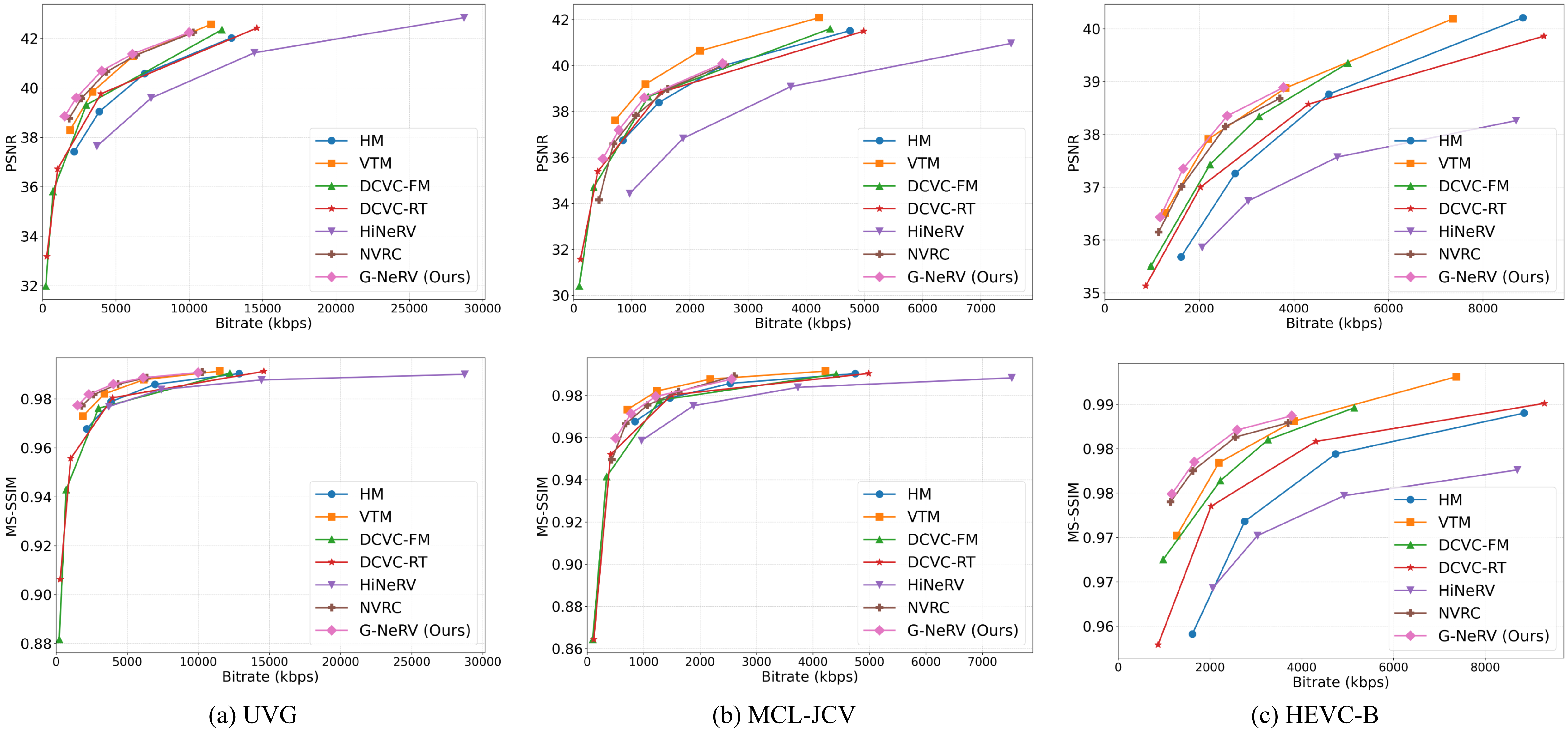}
    \caption{Rate-distortion curves on (a) UVG, (b) MCL-JCV and (c) HEVC B datasets.}
    \label{fig:RDCurve}
\end{figure*}

\subsection{Databases and Metrics}
We evaluate our video codec on three video datasets that are widely used in the neural video coding literature: UVG~\cite{UVG}, MCL-JCV~\cite{MCL-JCV}, and HEVC Class B~\cite{hevc}. Following the common practice \cite{Review_Gao}, video quality is measured using PSNR and MS-SSIM  \cite{ms-ssim} on YUV 4:2:0 colorspace~\cite{hevc}.
The PSNR is calculated by a weighted average on Y-, U-, V-channel with weights 6:1:1, respectively.
The MS-SSIM is only calculated on the Y-channel.
The coding performance of \name is estimated by calculating the BD-rate \cite{bd} values between our method and each baseline model (the latter is used as the anchor). The computational complexity figures in terms of encoding/decoding FPS, kMACs/px, and model parameters are also reported. Here, the same GPU is used for fairly measuring encoding/decoding FPS figures. For INR-based codecs, different complexity of the model will be used in different rate, so we report the average complexity.

\subsection{Baseline Methods}
We benchmark our approach against three video codec families covering the landscape of standardized video codecs (HEVC HM 18.0 \cite{HEVC_overview} and VVC VTM 20.0~\cite{VVC_overview}), autoencoder-based neural codecs (DCVC-FM~\cite{DCVC_FM} and DCVC-RT~\cite{DCVC_RT}), and INR-based codecs (HiNeRV~\cite{HiNeRV}, and NVRC~\cite{NVRC}). Notably, VTM, DCVC-RT and NVRC represent the state-of-the-art in each family.

\begin{table*}[t]
\centering
\caption{BD-rate results on three databases, with our codec as the anchor.}
\resizebox{\textwidth}{!}{
\begin{tabular}{r|rr|rr|rr|rr|c|c}
\toprule[1.2pt]
BD-rate (\%) & \multicolumn{2}{c|}{UVG} & \multicolumn{2}{c|}{MCL-JCV} & \multicolumn{2}{c|}{HEVC B} & \multicolumn{4}{c}{model complexity} \\
\midrule[1.1pt]
codec                    & $\text{PSNR}$ & $\text{MS-SSIM}$ & $\text{PSNR}$ & $\text{MS-SSIM}$ & $\text{PSNR}$ & $\text{MS-SSIM}$ & enc. FPS & dec. FPS & params (M) & kMACs/px \\
\midrule
HM 18.0 (RA)              & -43.80\%         & -49.44\%            & -18.90\%         & -20.30\%            & -37.86\%         & -40.21\%            & 0.06     & 39.5     & -           & - \\
VTM 20.0 (RA)             & -14.68\%         & -22.21\%            & 27.48\%         & 19.98\%            & -4.67\%          & -24.72\%            & 0.02     & 23.1     & -           & - \\
DCVC-FM                   & -25.28\%         & -45.25\%            & 0.18\%         & -16.36\%            & -20.98\%         & -34.40\%            & 5.24     & 5.45     & 16.13           & 859.28 \\
DCVC-RT                   & -32.73\%         & -44.32\%              & -5.34\%           & -8.06\%            & -30.40\%         & -49.93\%            & 28.4     & 113     & 18.08           & 165.53 \\
HiNeRV                      & -63.51\%          & -61.83\%             & -60.49\%         & -51.13\%             & -66.85\%         & -79.57\%            & 0.02     & 27.7     & 14.2           & 215.6 \\
NVRC                      & -8.86\%          & -7.93\%             & -10.24\%         & -10.41\%             & -12.04\%         & -13.46\%            & 0.01     & 16.5     & 16.8           & 582.1 \\\midrule
\name \textbf{(ours)}                      & 0.0\%           & 0.0\%            & 0.0\%         & 0.0\%            & 0.0\%         & 0.0\%            & 0.01     & 14.9     & 17.5           & 612.1 \\
\bottomrule[1.2pt]
\end{tabular}
}
\label{tab:bd-rate}
\end{table*}

\subsection{Overall Performance}
The BD-rate results between each baseline method (used as the anchor) and \name are shown in Table~\ref{tab:bd-rate}. It has been noted that \name achieves evident bitrate savings, with 14.68\%, 25.28\%, and 8.86\% BD-rate gains against VTM 20.0 (RA), DCVC-FM, and NVRC, respectively. Similar performance has also been observed on the HEVC B. The performance is inferior to the VTM on MCL-JCV, but still superior to that of INR-based codecs. Moreover, the complexity figures of \name and other baseline models have also been reported in Table~\ref{tab:bd-rate}, where \name increases the decoding time by 10.73\% compared to NVRC, along with a small model parameter overhead.

\begin{figure}[t]
    \centering
    \includegraphics[width=1.0\linewidth]{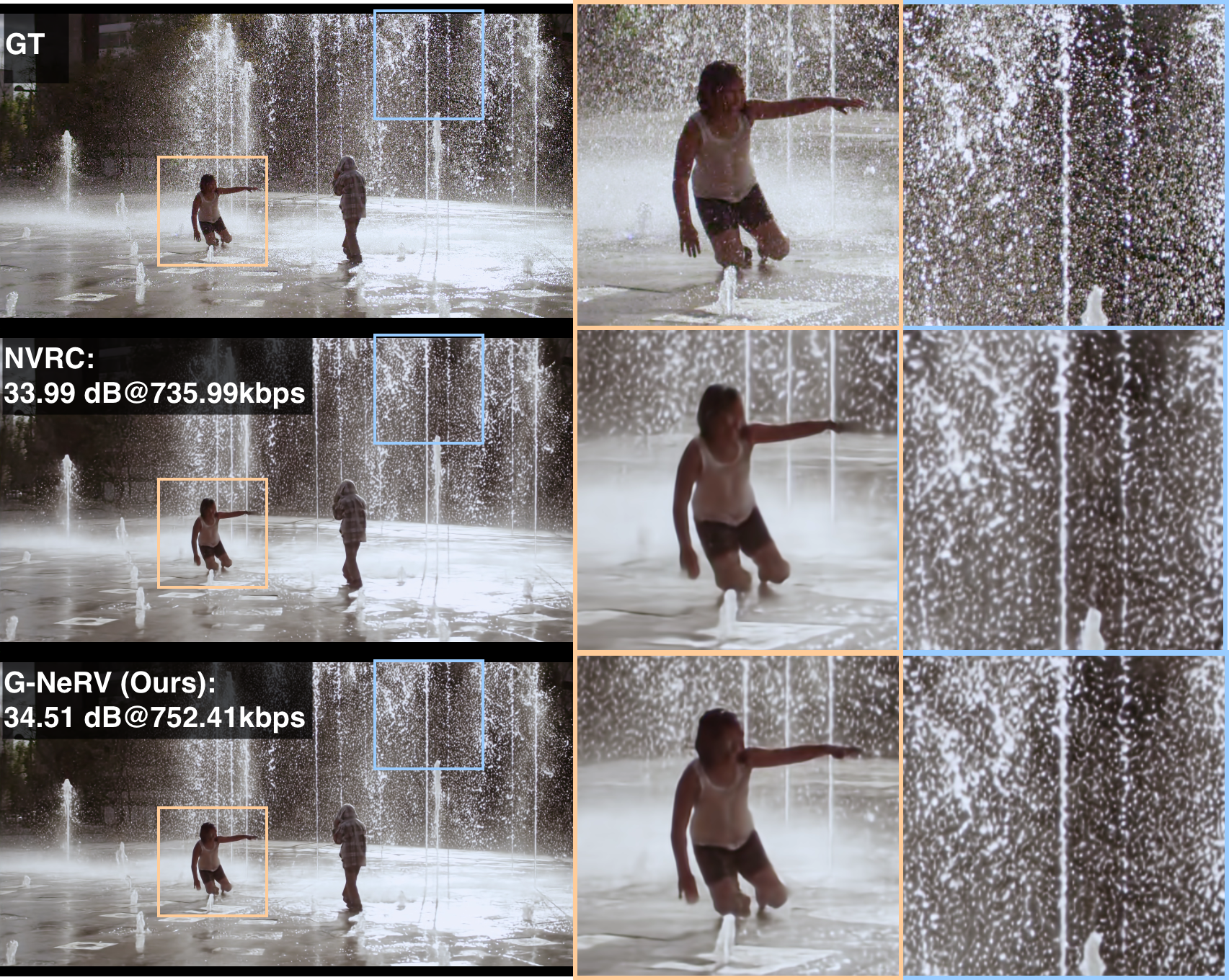}
    \caption{The qualitative comparison between GT, NVRC, and \name.}
    \label{fig:PerceptualVisualization}
\end{figure}

\subsection{Ablation Studies}\label{sec:ablation}
Each key component in our \name framework has been verified through ablation studies, with results shown in Table~\ref{tab:modules} and~\ref{tab:TemporalWindow}.
Here, the ablation experiments were conducted on the UVG dataset, with video quality measured by PSNR, and the original model employed as the anchor when calculating BD-rate results.

\paragraph{The effectiveness of the multi-scale 1D convolution layer} is evaluated by replacing it with a single-scale 1D convolution layer (V1.1)~\cite{1DConv}.

\paragraph{The effectiveness of the adaptive gate} is ablated by removing the gate (V2.1), using the channel-only gate (V2.2), or using the spatial-only gate (V2.3).

\paragraph{The effectiveness of temporal message passing} is evaluated by conducting experiments on removing temporal message passing (V3.1), using temporal message passing but with random neighborhoods (V3.2), adding temporal message passing only in grid-based encoding (V3.3), or adding temporal message passing only in decoder blocks (V3.4).

\paragraph{The effectiveness of the update strategy in decoding} is confirmed by using the reconstruction approach which fills the memory bank randomly (V4.1), rather than using the proposed update strategy in the decoding stage.

\paragraph{The effectiveness of the overall framework} is verified by instantiating an alternative temporal redundancy exploitation mechanism, DNeRV\cite{DNeRV}, for the NVRC. Specifically, we modified the representation model in NVRC by incorporating the mechanism described in DNeRV and kept the quantization and entropy models in NVRC.

\paragraph{The influence of the temporal radius} is investigated by conducting experiments with different radii $K$.

It can be observed from Table~\ref{tab:modules} that when compared to the original \name, all the variants V1.1-V4.1 achieve inferior performance with BD-rate losses or complexity increase. This confirms the effectiveness of each tested component. We cannot conduct ablative experiments with enabling message passing mechanism but removing memory bank since it requires large GPU memory and computation complexity when encoding HD video sequences. Moreover, when \name is integrated with a different INR model (V5.1), the BD-rate is still significant over the corresponding baseline, which indicates the generalization of the proposed method. Finally, Table~\ref{tab:TemporalWindow} reports the R-D performance, encoding, and decoding time of \name when different $K$ configurations are applied. It can be noted that the increase in $K$ does improve the R-D performance, with the compromise of encoding and decoding complexity. To trade off between performance and complexity, we empirically set $K=3$ in this work.

\begin{table}[t]
\caption{The ablation study results, with the original model as anchor.}
\vspace{-7pt}
\label{tab:modules}
\centering
\setlength{\tabcolsep}{1.5mm}{%
\begin{tabular}{ccccc}
\toprule
Version & BD-rate & Enc. FPS & Dec. FPS     & params (M) \\ \midrule
V1.1    & 1.94\%   & 0.009    & 14.9        & 17.4       \\ \midrule
V2.1    & 14.99\%  & 0.009    & 14.9        & 17.4 \\
V2.2    & 2.41\%   & 0.009    & 14.8        & 17.5      \\
V2.3    & 2.68\%   & 0.009    & 14.8        & 17.5      \\ \midrule
V3.1    & 8.89\%   & 0.010    & 16.5        & 16.8      \\ 
V3.2    & 9.40\%   & 0.010    & 15.9        & 17.5      \\
V3.3    & 2.01\%   & 0.010    & 15.9        & 17.0      \\
V3.4    & 3.27\%   & 0.009    & 14.8        & 17.3      \\ \midrule
V4.1    & 0.0\%    & 0.009    & 7.76        & 17.5      \\ \midrule
V5.1    & 7.91\%   & 0.005    & 15.7        & 16.9      \\ \midrule
Ours    & 0.0\%   & 0.009    & 14.9        & 17.5      \\
\bottomrule
\end{tabular}
}
\end{table}

\begin{table}[t]
\caption{The study for the neighborhood size $K$.}
\vspace{-7pt}
\label{tab:TemporalWindow}
\centering
\setlength{\tabcolsep}{2mm}{%
\begin{tabular}{ccccc}
\toprule
K & BD-rate    & Enc. FPS   & Dec. FPS  &kMACs/px \\ \midrule
1 & 1.9\%      & 0.012      & 15.9      & 601.2\\
2 & 1.3\%      & 0.010      & 15.4      & 605.8\\
3 (Ours) & 0.0\%      & 0.009      & 14.8      & 612.1\\
4 & -0.3\%     & 0.008      & 14.1      & 620.9\\
5 & -0.5\%     & 0.006      & 13.4      & 629.2\\ \bottomrule
\end{tabular}
}
\end{table}

\section{Conclusion}
This paper proposes a novel graph-inspired neural representation framework for video compression, \name,  to explicitly exploit the temporal redundancy in video sequences.
\name employs newly designed message passing modules that combine an adaptive gate to share information between latent embeddings. Each message passing module is applied together with a memory bank that dynamically generates and stores neighbor embeddings, which is motivated by the reference frame buffer in conventional video codecs. The experimental results show that this new framework can evidently improve overall coding performance by 8.86\% in BD-rate savings measured by PSNR on UVG dataset, when it is integrated with an advanced neural representation compression framework for the video coding task.

\paragraph{Impact of this work.}
The proposed graph-inspired INR framework can potentially (although it has not been fully tested) be applied to existing INR-based video codecs to improve coding performance through enhanced temporal redundancy exploitation. The proposed message passing module can also be adapted for other applications, such as INR-based dynamic 3D scene reconstruction and compression.

\paragraph{Limitations and future work.} For video compression, one of the primary limitations of this work is the latency issue. Similar to other INR-based methods, \name requires seeing an entire video sequence (or a large number of video frames) during the encoding process in order to maintain coding performance comparable to that of conventional video codecs. This has been investigated in some existing models \cite{PNVC}, but it will remain as the future work.

\paragraph{Acknowledgement}
This work was supported by the funding from EPSRC DLA Award, the University of Bristol, ByteDance US, and the UKRI MyWorld Strength in Places Programme (SIPF00006/1).

\bibliography{aaai2027}

\section{Appendix}

\subsection{Pseudo-code for \name}

We present the pesudo-code for the encoding and decoding of our \name pipeline in Algorithm~\ref{alg:encoding} and \ref{alg:decoding}. The notations in these two algorithms are the same as those in the main paper. Specifically, $h_t^{(\ell)}$ denotes the embedding with index $t$ at layer $\ell$. $\mathbf{H}^{\text{grid}}_t$ and $\mathbf{H}^{(\ell)}_t$ represent the temporal neighborhood with the center index $t$ in the grid encoding and decoding layer $\ell$, respectively. $K$ is the radius of the neighborhood. $p(\cdot)$ stands for the up-sampling block in the decoding layer. $\mathcal{M}$ denotes the memory bank, and $\mathcal{D}$ represents the down-sampling module in the memory bank. Since our codec incorporate the learnable parameters in the quantization and entropy model, we use $\phi$, and $\psi$ to denotes these parameters in the codec.

\subsection{Implementation Details}
\subsubsection{\name Configuration}
In \name, we employ four configurations with different initial representation sizes to support multiple rates. These configurations differ only in their latent-space dimensions. Detailed settings are provided in Table~\ref{aptab:implementation_details}, where specific hyper-parameters for our work are highlighted in gray and hyper-parameters shared with NVRC are left unshaded. The quantization and entropy-model configurations follow those of NVRC~\cite{NVRC}.

\subsubsection{Training Details}
During the training (encoding) stage, without losing generality, we train the model by predicting the patches for the frame, following HiNeRV~\cite{HiNeRV} to reduce GPU memory consumption. Specifically, each frame is divided into $120 \times 120$ patches, resulting in 144 patches per frame. For each patch, its temporal neighborhood is constructed using the spatially corresponding patches from neighboring frames. When sampling from and updating the memory bank, we operate only on the patch corresponding to the embedding at temporal index $t$. During the decoding stage, we perform frame-wise inference to avoid boundary artifacts introduced by patch-wise inference. To ensure consistency between patch-wise training and frame-wise decoding, the embeddings are appropriately padded during training, following the strategy described in HiNeRV~\cite{HiNeRV}.

\begin{algorithm}[t]\small
  \caption{Encoding process}
  \label{alg:encoding}
  \textbf{Input}: Video sequence index $\{t\}_{t=1}^{T}$.\\
  \textbf{Parameter}: Learnable parameters $\theta$, $\phi$, and $\psi$; window radius $K$.\\
  \textbf{Output}: Bitstream $\mathcal{B}$.
  \begin{algorithmic}[1]
    \WHILE{not converged}
      \STATE Sample a frame index $t$. 
      \quad \green{\# Frame sampling}
      \STATE $\textbf{H}_t^{\text{grid}}\leftarrow\operatorname{Grid\_Enc}(t,K)$.
      \quad \green{\# Grid encoding}
      \STATE $\widetilde{h}_t^{(0)}\leftarrow g^{(0)}(\textbf{H}_t^{\text{grid}})$.
      \FOR{each layer $\ell\in [1,L]$}
        \STATE $h_t^{(\ell)}
        \leftarrow p(\widetilde{h}_t^{(\ell-1)})$.
        \quad \green{\# Pre-message embedding}
        \STATE $\mathcal{M}^{(\ell)}_t\leftarrow\mathcal{D}^{(\ell)}(h_t^{(\ell)})$ \quad \green{\# Memory bank update}
        \STATE \green{\# Neighborhood construction}
        \STATE $\mathbf{H}_t^{(\ell)}
        \leftarrow \text{Construct}(\mathcal{M}^{(\ell)},h_t^{(\ell)})$.
        \STATE $\widetilde{h}_t^{(\ell)}\leftarrow g^{(\ell)}\left(\mathbf{H}_t^{(\ell)}\right)$.\quad \green{\# Message passing}
      \ENDFOR
      \STATE $\hat{x}_t\leftarrow
      \operatorname{Output\_head}(\widetilde{h}_t^{(L)})$.
      ~\green{\# Frame reconstruction}
      \STATE $\mathcal{L}\leftarrow R+\lambda D$.
      \qquad\qquad \green{\# Rate-distortion loss}
      \STATE Update $\theta$, $\phi$, and $\psi$ using $\mathcal{L}$.
    \ENDWHILE
    \STATE Quantize and entropy-code $(\theta,\phi,\psi)$ into
    $\mathcal{B}$.
    \RETURN $\mathcal{B}$.
  \end{algorithmic}
\end{algorithm}

For training YUV 420 data, we first up-sample the U- and V-channels to the same size as the Y-channel and feed them to the model. The outputs of the U- and V-channels are then down-sampled to the original resolution and compared to the ground-truth.
The distortion in the training loss is a combination of $\ell_2$ and MS-SSIM loss. The $\ell_2$ loss is formulated as:
\begin{equation}
\begin{aligned}
    d_{\ell2}=&\frac{6}{8}\times\sum_{t=1}^T\|x_t^Y-\hat{x}_t^Y\|_2^2\\
    +&\frac{1}{8}\times\sum_{t=1}^T\|x_t^U-\hat{x}_t^U\|_2^2+\frac{1}{8}\times\sum_{t=1}^T\|x_t^V-\hat{x}_t^V\|_2^2,
\end{aligned}
\end{equation}
where $\hat{x}_t^Y$, $\hat{x}_t^U$, and $\hat{x}_t^V$ denote the output of the Y-, U-, and V-channels, respectively.

\begin{algorithm}[ht]\small
  \caption{Decoding process}
  \label{alg:decoding}
  \textbf{Input}: Bitstream $\mathcal{B}$\\
  \textbf{Parameter}: Window radius $K$.\\
  \textbf{Output}: Video sequence $\{\hat{x}_t\}_{t=1}^{T}$
  \begin{algorithmic}[1]
    \STATE $\mathcal{B}\leftarrow(\hat{\theta},\hat{\phi},\hat{\psi})$. \qquad\qquad\qquad \green{\# Entropy-decode}
    \STATE $f_{\hat{\theta}}\leftarrow\hat{\theta}$\qquad\qquad\qquad\qquad\quad \green{\# Network Reconstruction}
    \STATE \green{\# Index-order sweeping from first frame to last frame}
    \FOR{$t=1$ to $T$}
      \STATE $\textbf{H}_t^{\text{grid}}\leftarrow\operatorname{Grid\_Enc}(t,K)$.
      \quad \green{\# Grid encoding}
      \STATE $\widetilde{h}_t^{(0)}\leftarrow g^{(0)}(\textbf{H}_t^{\text{grid}})$.
      \FOR{each layer $\ell$}
        \FOR{each missing index
        $i$ in bank}
          \STATE $h_i^{(\ell)}
          \leftarrow p^{(\ell)}
          (\widetilde{h}_i^{(\ell-1)})$.
          \STATE $\mathcal{M}^{(\ell)}_t\leftarrow\mathcal{D}^{(\ell)}(h_i^{(\ell)})$.
          \STATE $\mathcal{Q}^{(\ell)}\leftarrow h_i^{(\ell)}$\qquad\qquad \green{\# Enqueue}
        \ENDFOR
        \STATE $h_t^{(\ell)}
        \leftarrow\operatorname{Pop}(\mathcal{Q}^{(\ell)},t)$.
        \qquad \green{\# Dequeue}
        \STATE \green{\# Neighbor construction}
        \STATE $\mathbf{H}_t^{(\ell)}\leftarrow \text{Construct}(\mathcal{M}^{(\ell)},h_t^{(\ell)})$.
        \STATE $\widetilde{h}_t^{(\ell)}\leftarrow g^{(\ell)}\left(\mathbf{H}_t^{(\ell)}\right)$.
        \qquad \green{\# Message passing}
      \ENDFOR
      \STATE $\hat{x}_t\leftarrow
      \operatorname{Output\_head}(\widetilde{h}_t^{(L)})$.
      ~\green{\# Frame reconstruction}
    \ENDFOR
    \RETURN $\{\hat{x}_t\}_{t=1}^{T}$.
  \end{algorithmic}
\end{algorithm}

The MS-SSIM loss is only computed on the Y-channel with a window size of $5$ and is formulated as:
\begin{equation}
    d_{\text{MS-SSIM}}=\sum_{t=1}^T\left(1-\text{MS-SSIM}(x_t^Y,\hat{x}_t^Y)\right)
\end{equation}

The total loss is a weighted combination of $d_{\ell2}$ and $d_{\text{MS-SSIM}}$, formulated as:
\begin{equation}
    d=d_{\ell2}+0.1\times d_{\text{MS-SSIM}},
\end{equation}
which aligns with both the commonly used PSNR-YUV (6:1:1) and MS-SSIM (Y) metrics. While we did not thoroughly study the weighting between the two terms, we found that this ratio offers both good PSNR and MS-SSIM performance.

For optimization, we train our codec in two stages using Adam~\cite{Adam} as the optimizer. The learning rates for the first and second stages are $2 \times 10^{-3}$ and $1 \times 10^{-4}$, respectively. In the first stage, we employ soft rounding and Kumaraswamy noise~\cite{C3} to enable gradient-based optimization through quantization. The soft-rounding temperature is annealed from $0.5$ to $0.3$, while the noise-scale ratio is reduced from $2.0$ to $1.0$. In the second stage, we apply Quant-Noise~\cite{Quant-Noise} with a ratio that increases from $0.5$ to $1.0$. For the UVG and HEVC datasets, the first and second stages are trained for 400 and 20 epochs, respectively. For the MCL-JCV dataset, we double the number of epochs to maintain a comparable encoding time. Moreover, $R$ is optimized once for every $8$ optimization steps of $D$.

\begin{figure}[ht]
    \centering
    \includegraphics[width=1.0\linewidth]{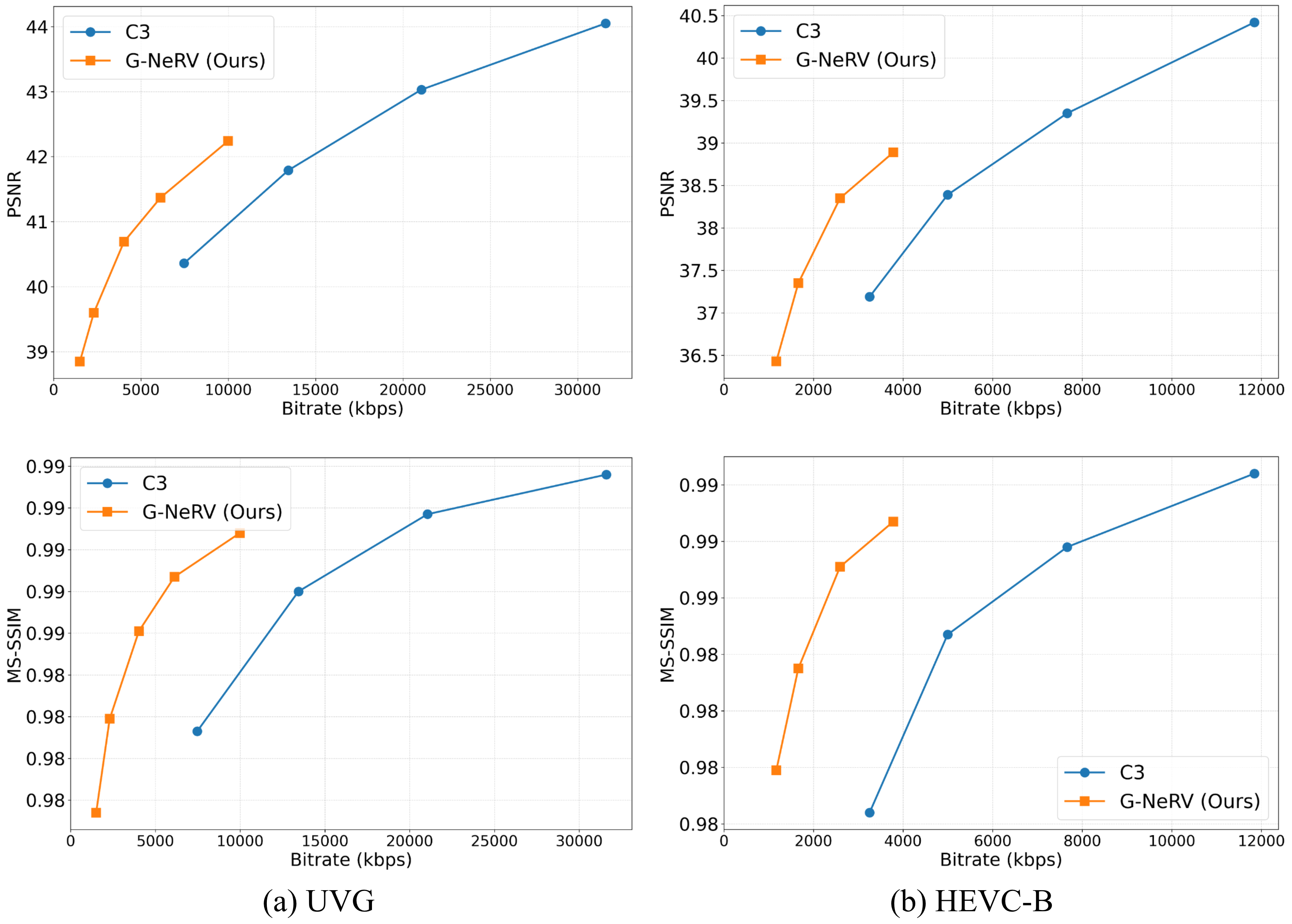}
    \caption{Rate-distortion curves on (a) UVG, and (b) HEVC B datasets.}
    \label{apfig:RDCurve}
\end{figure}

\subsubsection{Computing Infrastructure}
We use the high-performance GPU for all experiments. For measuring the complexity, which is conducted by the same high-performance GPU.

\begin{table*}[t]
\centering
\caption{Implementation details, specific hyper-parameters for our work are highlighted in gray.}
\resizebox{\textwidth}{!}{
\begin{tabular}{lcccc}
\toprule[1.2pt]
                           & Rate 1                      & Rate 2                      & Rate 3                       & Rate 4                       \\ \midrule
Lambda                     & 4                           & 64                          & 256                          & 1024                         \\
Channel dimensions         & {[}336, 168, 84, 42{]}      & {[}336, 168, 84, 42{]}      & {[}512, 256, 128, 64{]}      & {[}768, 384, 192, 96{]}      \\
Convolution kernel size    & \multicolumn{4}{c}{3}                                                                                                   \\
Expansion ratios           & \multicolumn{4}{c}{(3, 2, 2, 2)}                                                                                        \\
Grid temporal dimension    & \multicolumn{4}{c}{{[}200, 100, 50, 25{]} for UVG and HEVC, {[}50, 25, 12, 6{]} for MCL-JCV}                            \\
Grid spatio dimension      & {[}2, 4, 8, 16{]}           & {[}4, 8, 16, 32{]}          & {[}4, 8, 16, 32{]}           & {[}8, 16, 32, 64{]}          \\
Local grid temp. dimension & \multicolumn{4}{c}{{[}600, 300, 150{]} for UVG and HEVC, {[}150, 75, 38{]} for MCL-JCV}                                 \\
Local grid spa. dimension  & {[}8, 16, 32{]}             & {[}8, 16, 32{]}             & {[}16, 32, 64{]}             & {[}32, 64, 128{]}            \\ \midrule
\rowcolor{gray!20}
Neighborhood size          & \multicolumn{4}{c}{3}                                                                                                   \\
\rowcolor{gray!20}
1D convolution kernel size & \multicolumn{4}{c}{{[}1, 3, 5, 7{]}}                                                                                    \\
\rowcolor{gray!20}
Gate dimension             & {[}336, 336, 168, 84, 42{]} & {[}336, 336, 168, 84, 42{]} & {[}512, 512, 256, 128, 64{]} & {[}768, 768, 384, 192, 96{]} \\
\rowcolor{gray!20}
Bank channel ratios        & \multicolumn{4}{c}{(0.5, 0.5, 0.5, 0.5)}                                                                                \\
\rowcolor{gray!20}
Bank  spatio ratios        & \multicolumn{4}{c}{(0.25, 0.25, 0.25, 0.25)}                                                                            \\
\bottomrule[1.2pt]
\end{tabular}
}
\label{aptab:implementation_details}
\end{table*}

\begin{figure*}[t]
    \centering
    \includegraphics[width=1.0\linewidth]{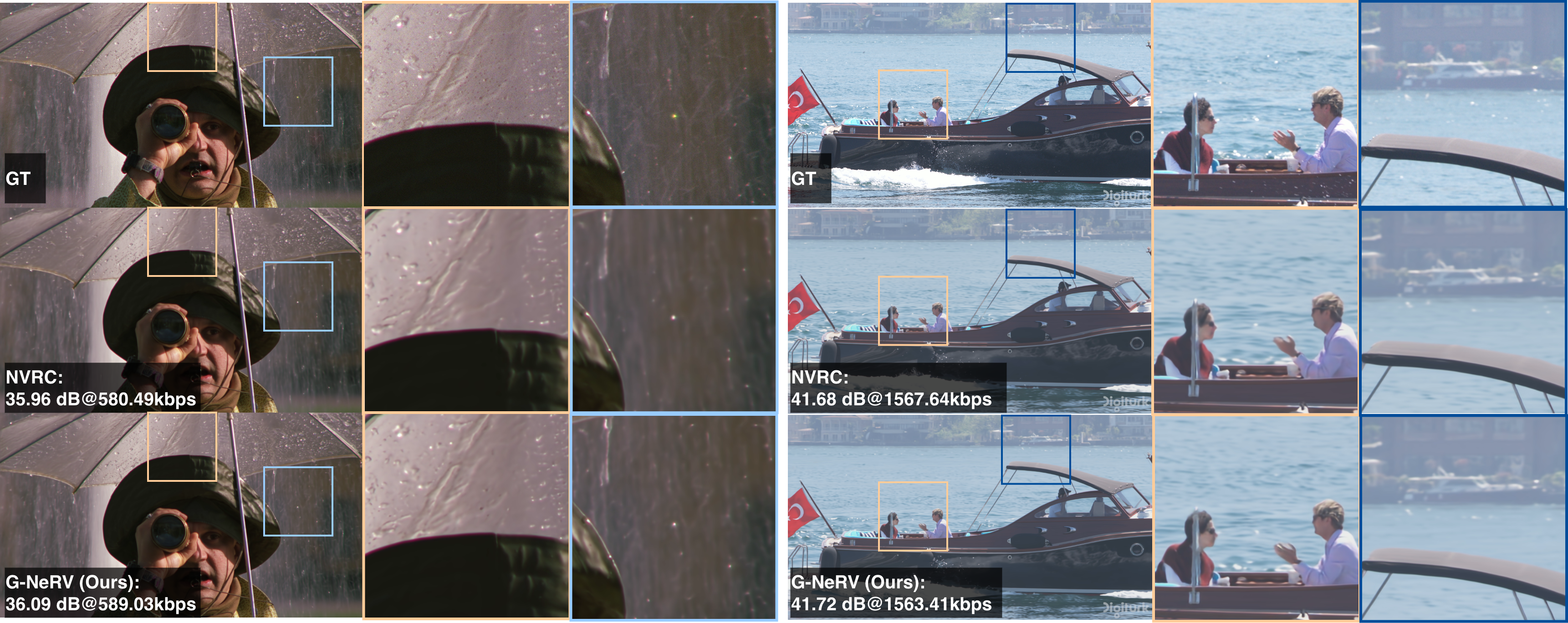}
    \caption{The qualitative comparison between GT, NVRC, and \name.}
    \label{ap_fig:PerceptualVisualization}
\end{figure*}

\subsubsection{More Experimental Results}
We further compare \name with C3~\cite{C3} on the UVG~\cite{UVG} and HEVC-B~\cite{hevc} datasets using the same experimental setup as in the main paper. The quantitative results are reported in Table~\ref{aptab:bd-rate}, and the corresponding rate-distortion curves are presented in Figure~\ref{apfig:RDCurve}. The results show that \name achieves substantial PSNR and MS-SSIM improvements over C3 at comparable bit rates, although increasing the decoding complexity. We omit results on the MCL-JCV dataset~\cite{MCL-JCV} because the quality range achieved using the original C3 configuration does not overlap with that of our codec, preventing a meaningful comparison.

Additional qualitative comparisons between the ground-truth frames and the reconstructions produced by NVRC and \name are presented in the Figure~\ref{ap_fig:PerceptualVisualization}. These results demonstrate the superior reconstruction quality achieved by \name.

\begin{table*}[ht]
\centering
\caption{BD-rate results on two databases, with our codec as the anchor.}
\resizebox{\textwidth}{!}{
\begin{tabular}{r|rr|rr|rr|c|c}
\toprule[1.2pt]
BD-rate (\%) & \multicolumn{2}{c|}{UVG} & \multicolumn{2}{c|}{HEVC B} & \multicolumn{4}{c}{model complexity} \\
\midrule[1.1pt]
codec                    & $\text{PSNR}$ & $\text{MS-SSIM}$ & $\text{PSNR}$ & $\text{MS-SSIM}$ & enc. FPS & dec. FPS & params (M) & kMACs/px \\
\midrule
C3                      & -47.14\%          & -63.03\%              & -48.19\%         & -62.79\%            & 3.21     & 17.5     & 0.01           & 3.7 \\\midrule
\name \textbf{(ours)}                      & 0.0\%           & 0.0\%                & 0.0\%         & 0.0\%            & 0.01     & 14.9     & 17.5           & 612.1 \\
\bottomrule[1.2pt]
\end{tabular}
}
\label{aptab:bd-rate}
\end{table*}

\end{document}